\pdfoutput=1

\documentclass[11pt]{article}

\usepackage{ACL2023}

\usepackage{times}
\usepackage{latexsym}
\usepackage{graphicx}
\usepackage{xcolor}
\usepackage{linguex}
\usepackage{booktabs}

\usepackage[T1]{fontenc}

\usepackage[utf8]{inputenc}

\usepackage{microtype}

\usepackage{inconsolata}

\definecolor{plum}{HTML}{92268F}

\title{How to Plant Trees \includegraphics[height=0.5cm]{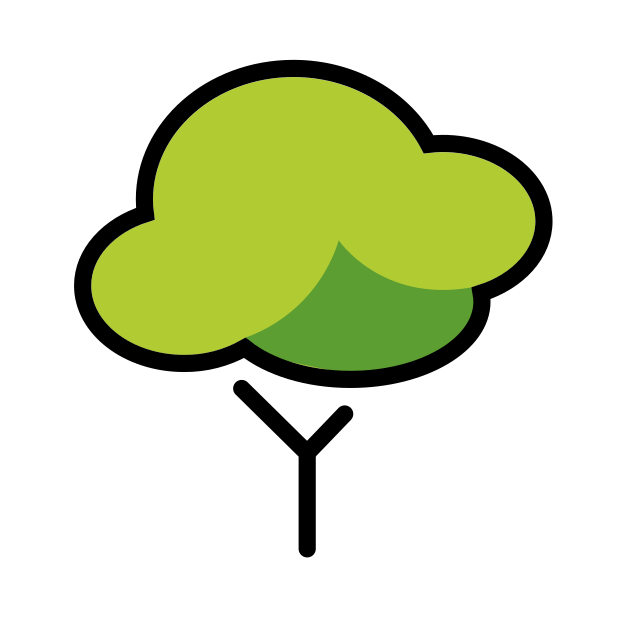} in LMs: Data and Architectural\\Effects on the Emergence of Syntactic Inductive Biases}

\author{Aaron Mueller \\
  Johns Hopkins University\\
  \texttt{amueller@jhu.edu} \\\And
  Tal Linzen \\
  New York University \\
  \texttt{linzen@nyu.edu} \\}

\begin{document}
\maketitle
\begin{abstract}
Accurate syntactic representations are essential for robust generalization in natural language. Recent work has found that pre-training can teach language models to rely on hierarchical syntactic features---as opposed to incorrect linear features---when performing tasks after fine-tuning. We test what aspects of pre-training are important for endowing encoder-decoder Transformers with an inductive bias that favors hierarchical syntactic generalizations. We focus on architectural features (depth, width, and number of parameters), as well as the genre and size of the pre-training corpus, diagnosing inductive biases using two syntactic transformation tasks: question formation and passivization, both in English. We find that the number of parameters alone does not explain hierarchical generalization: model depth plays greater role than model width. We also find that pre-training on simpler language, such as child-directed speech, induces a hierarchical bias using an order-of-magnitude less data than pre-training on more typical datasets based on web text or Wikipedia; this suggests that in cognitively plausible language acquisition settings, neural language models may be more data-efficient than previously thought.
\end{abstract}

\section{Introduction}
Accurate syntactic representations are necessary for robust generalization to new natural language inputs and for the generation of correct outputs. Consider the problem of identifying the subject of ``said'' in the following sentence:

\ex. Can \textbf{\textcolor{red}{you}} repeat what the \textbf{\textcolor{blue}{senator}} next to the \textbf{\textcolor{red}{cats}} \textbf{\textcolor{blue}{said}}?\label{ex:heuristics}

Typical language models (LMs), which receive linear sequences of words as input, could conceivably rely on a linear or positional feature that usually, but does not always, identifies the correct subject of a verb. An LM could learn, for example, that the first noun in the sentence is always the subject. This heuristic works for many simple sentences, but fails in Ex.~\ref{ex:heuristics}: here, the first noun is ``\textcolor{red}{\textbf{you}}'', and so this heuristic would lead the LM to incorrectly interpret the sentence as meaning ``Can you repeat what \textcolor{red}{\textbf{you}} said?'' The LM could also learn that the subject of the verb is the noun closest to the verb in the linear order of the sentence, in which case it would interpret Ex.~\ref{ex:heuristics} as ``Can you repeat what the \textbf{\textcolor{red}{cats}} said?'' By contrast, an LM that represents the sentence as hierarchically structured will correctly identify \textbf{\textcolor{blue}{senator}} as the subject of the embedded clause that contains the verb \textbf{\textcolor{blue}{said}}. This example demonstrates that a preference for syntactic features over linear features is \emph{required} for robust linguistic generalization.

The success of large-scale pre-training across fine-tuning tasks suggests that exposure to natural language may teach models to rely on appropriate syntactic features instead of heuristics (even though models still often rely on heuristics; \citealt{mccoy-etal-2019-right}). This hypothesis is supported by the finding that, given minimal pairs of grammatical and ungrammatical sentences, the probability distribution over sentences defined by LMs often favors the grammatical sentence \citep{marvin-linzen-2018-targeted,hu-etal-2020-systematic}. A related line of work has shown that, through pre-training, LMs can under some circumstances acquire syntactic inductive biases which are then applied to fine-tuning tasks, whereas models which have not been pre-trained do not have such inductive biases (\citealt{warstadt-bowman-2020-linguistic,warstadt-etal-2020-learning,lovering2021predicting,mueller-etal-2022-coloring}).

When does pre-training endow LMs with a syntactic inductive bias? In this study, we address two specific sub-questions: (1) Which architectural features make a syntactic inductive bias more likely to emerge in a Transformer LM? (2) How is the inductive bias affected by the genre and size of the pre-training corpus? We investigate these questions by evaluating a range of Transformer encoder-decoder models based on T5 \citep{raffel-etal-2020-t5}. We evaluate both existing publicly available models and models that we pre-train ourselves; we explore a variety of model widths (embedding and hidden dimension, feed-forward layer size) and depths (number of layers), and pre-train on corpora of varying genres and sizes. We then evaluate models' inductive biases by observing their out-of-distribution generalization when fine-tuned on syntactic transformations tasks (\S\ref{sec:methods}). We find that \textbf{depth matters more than width for the acquisition of hierarchical biases} (\S\ref{sec:arch-effects}), and that \textbf{pre-training on simpler language induces hierarchical biases using far less data} (\S\ref{sec:style} and \S\ref{sec:data-quantity}). This last finding suggests that in language acquisition settings in which the training corpus  more closely resembles the language that children are exposed to, Transformers may be more sample-efficient than previously thought.

Our code is available on GitHub.\footnote{\url{https://github.com/aaronmueller/emergent-syntax}}

\section{Background and Motivation}
\label{sec:background}

Every finite training set is consistent with multiple generalizations. We use the term \textbf{inductive bias} to refer to the set of assumptions that a model relies on when generalizing to new data. Our usage includes any factor that leads the model to generalize in one way rather than another \cite{mitchell1980need}; this can include not only the model's architecture, but also representations learned from prior or concurrent training on tasks that are related to the target task \cite{caruana-1997-multitask}, and in particular self-supervised pre-training \cite{lovering2021predicting}.

We can infer a model's inductive bias by observing how it generalizes out of distribution after training on a dataset that is compatible with multiple generalizations. Applying this methodology, \citet{mccoy-etal-2018-poverty}, \citet{mccoy-2020-syntax-trees}, and \citet{petty-frank-2021-transformers} find that LSTM and Transformer encoder-decoder models trained from scratch (without pre-training) on syntactic transformations, such as converting a declarative sentence into a question (\S\ref{sec:syntactic-transformations}), do not generalize in a hierarchical manner. By contrast, \citet{mueller-etal-2022-coloring} find that certain \textbf{pre-trained} encoder-decoder models---including T5 and BART \citep{lewis-etal-2020-bart}---\emph{do} generalize hierarchically after fine-tuning. 
\citet{warstadt-bowman-2020-linguistic} and \citet{warstadt-etal-2020-learning} report similar results for the pre-trained masked LM RoBERTa \citep{liu-etal-2019-roberta}, though in their study a robust syntactic inductive bias only emerged when the training corpus was much larger than a human might be exposed to.

Previous work on the effect of training corpus size and genre on syntactic generalization includes \citet{huebner-etal-2021-babyberta}, who find that masked LMs show stronger syntactic abilities after training on a few million words of child-directed speech than a similar amount of Wikipedia or news text; they do not, however, explore whether similar abilities arise from training on a larger amount of Wikipedia text. 
\Citet{van-schijndel-etal-2019-quantity} report experimental results suggesting that scaling the training corpus or model size is unlikely to result in human-like syntactic abilities for LSTM LMs, but they only vary model width and only train on Wikipedia data. We fill the gap between these studies by investigating the influence of multiple component of the Transformer architecture and by training on corpora of varying genres and sizes. 

Our work is related more broadly to the syntactic LM evaluation literature. In this style of work, evaluation is typically performed using minimal pairs, where a grammatical and ungrammatical sentence or completion are provided to a model, and the model is expected to assign a higher probability to the grammatical variant. 
Syntactic evaluations have found that LSTM- \citep{hochreiter-1997-lstm} and Transformer-based \citep{vaswani-2017-attention} LMs are sensitive to grammatical number and gender in subject-verb agreement and reflexives \cite{hu-etal-2020-systematic,marvin-linzen-2018-targeted,goldberg-syntax-2019,lakretz-etal-2021-recursive,gauthier-etal-2020-syntaxgym}. LMs are also sensitive to filler-gap dependencies \cite{wilcox-etal-2018-rnn} and, to a lesser, extent, negative polarity items \cite{marvin-linzen-2018-targeted,warstadt-etal-2020-blimp-benchmark}. This holds across languages \cite{mueller-etal-2020-clams,ravfogel-etal-2018-lstm} and across grammatical/typological features \cite{ravfogel-etal-2019-studying}.

\begin{figure*}
    \centering
    \includegraphics[width=0.98\linewidth]{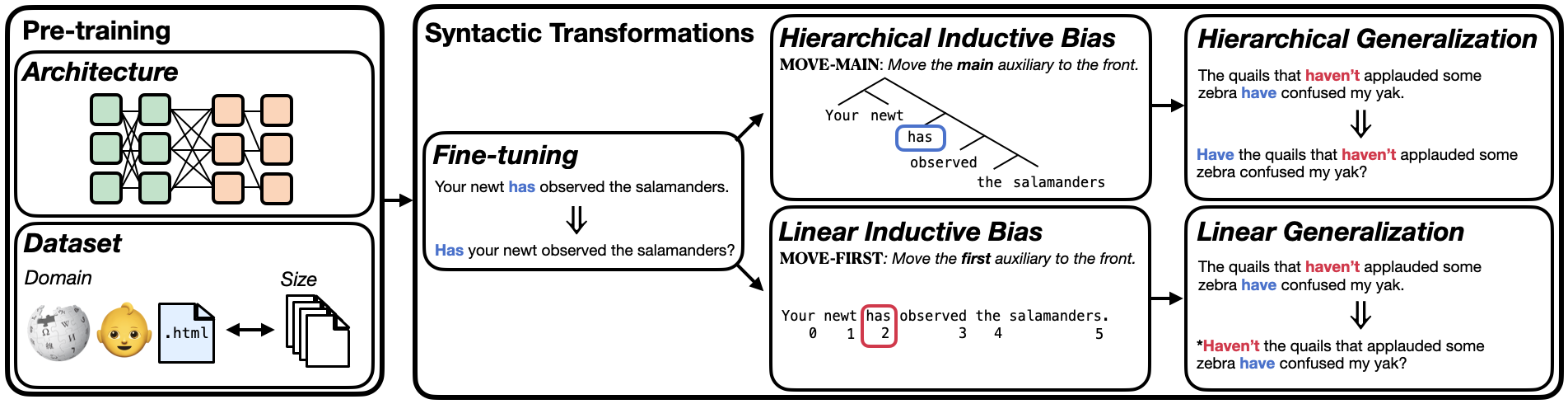}
    \caption{The syntactic transformations paradigm. A pre-trained model is fine-tuned on examples that are consistent with both syntactic (hierarchical) and positional/word order (linear) explanations. Then, it is evaluated on examples where only a model with a syntactic inductive bias will produce the correct output. We investigate which components of pre-training induce hierarchical inductive biases. Adapted from \citet{warstadt-etal-2020-learning} and \citet{mueller-etal-2022-coloring}.}
    \label{fig:transformations}
\end{figure*}

Overall, prior work has shown that pre-training can impart hierarchical inductive biases to LMs. The goal of this study is to examine which aspects of pre-training---specifically, architecture and training data---contribute to the emergence of this bias.

\section{Syntactic Transformations}\label{sec:syntactic-transformations}
To evaluate the linguistic inductive biases of our models, we employ the \emph{poverty of the stimulus} design \cite{wilson-2006-learning}: we fine-tune a model on ambiguous data and then evaluate it on out-of-distribution examples where only the desired inductive bias will result in the correct outputs. Here, we use the \textbf{syntactic transformations} paradigm \cite{frank-mathis-2007-transformational} summarized in Figure~\ref{fig:transformations}, and observe whether models generalize according to hierarchical linguistic rules or according to surface heuristics based on word position or relative word ordering. We evaluate on English question formation and passivization, using the English datasets of \citet{mueller-etal-2022-coloring} (themselves based on \citealt{mccoy-2020-syntax-trees}).

\subsection{Question Formation}

Here, the task is to transform a declarative sentence into a polar yes/no question by moving the auxiliary verb to the start of the sentence. The competing hypotheses are \textsc{Move-First} and \textsc{Move-Main} (see Figure~\ref{fig:transformations} for examples). We train the models on sentences that are consistent with both hypotheses, where the main auxiliary is always the linearly first auxiliary in the input sentence. Then, in the generalization examples, we append a relative clause (RC) to the subject, such that the main auxiliary is now the linearly \emph{second} auxiliary in the input. A model that acquired \textsc{Move-Main}---that is, one that has a hierarchical inductive bias---will correctly identify the main auxiliary verb and move it to the front, meaning that it should still produce the correct output. A model that learned \textsc{Move-First} will move the first auxiliary to the front, resulting in ungrammatical outputs (Figure~\ref{fig:transformations}).

\setlength{\Exlabelwidth}{0.25em}
\setlength{\SubExleftmargin}{1.35em}

\subsection{Passivization}

In this task, the goal is to transform an active sentence into a passive one. This requires various insertions, deletions, reinflections, and movements, making this task a potentially more difficult one than question formation. Here, we evaluate the movement of the object to the subject position. The competing hypotheses here are \textsc{Move-Second} and \textsc{Move-Main}. We train the models on sentences where the object is always the linearly second noun in the sentence. Then, in the generalization examples, we append a prepositional phrase (PP) to the subject, such that the object is now the linearly \emph{third} noun. If a model acquires the generalization \textsc{Move-Main} (consistent with a hierarchical inductive bias), it will detect the object and move it to the front, producing the correct output. If it acquires \textsc{Move-Second}, it will move the linearly second noun phrase even in the generalization examples (where, again, the correct noun to move is actually the linearly third one), and as such will output ungrammatical sequences. For example:

\ex.\textit{Passivization}
    \a. \textit{Training}: The raven observed the \textbf{\textcolor{plum}{newts}} (near the yak). $\Rightarrow$ The \textbf{\textcolor{plum}{newts}} (near the yak) were observed by the raven.
    \b. \textit{Generalization}: The salamander behind the \textbf{\textcolor{red}{ravens}} applauded the \textbf{\textcolor{blue}{peacock}}. $\Rightarrow$ ?
    \c. \textsc{Move-Main} (correct): The \textbf{\textcolor{blue}{peacock}} was applauded by the salamander behind the \textbf{\textcolor{red}{ravens}}.
    \d. \textsc{Move-Second} (incorrect): The \textbf{\textcolor{red}{ravens}} were applauded by the salamander.

\subsection{Evaluation Metrics}
For both syntactic transformations, we evaluate models' outputs using two metrics. The first is \textbf{\textcolor{blue}{sequence accuracy}}, which measures the percentage of inputs for which the model's full output sequence is exactly correct. This is a strict metric that does not capture solely the syntactic phenomenon we investigate, but also penalizes the model for other errors, such as word substitution errors. We also report more targeted metrics for each of the tasks: \textbf{\textcolor{orange}{main auxiliary accuracy}} for question formation, which measures how often the first word of the output sentence is the main auxiliary; and \textbf{\textcolor{orange}{object accuracy}} for passivization, which measures how often the noun that gets moved to the start of the sentence is the object.

\section{Overview of Experimental Paradigm}\label{sec:methods}
All of our experiments involve fine-tuning variants of T5, a Transformer encoder-decoder model pre-trained using a span denoising objective: contiguous token sequences are masked in the input sequence and then reconstructed in the output sequence. We either use the publicly available pre-trained ``efficient'' T5 models released by \citet{tay-etal-2022-scale},\footnote{The term ``efficient'' here contrasts  the models of \citet{tay-etal-2022-scale} with the original T5 models of \citet{raffel-etal-2020-t5}, which are Pareto-inefficient with respect to downstream performance and number of parameters. The ``efficient'' models achieve similar performance across tasks using fewer parameters by using a deeper (more layers) and narrower (smaller hidden size/feed-forward size) architecture.\label{footnote:efficient}} or pre-train models ourselves using the \texttt{transformers} library \cite{wolf-etal-2020-transformers}.

The syntactic transformation datasets we fine-tune on are the English datasets of \citet{mueller-etal-2022-coloring}, which consist of 100,000 training examples; 10,000 in-distribution test examples, which test whether the models have learned the task; and 10,000 out-of-distribution generalization examples, which reveal models' inductive biases. 

We adopt \citeauthor{mueller-etal-2022-coloring}'s hyperparameters (App.~\ref{app:hyperparameters}). We fine-tune for 10 epochs (approximately 7500 training steps), and every 500 steps we save a checkpoint and evaluate it. Across models, accuracy on the in-distribution test set generally reaches 100\% within 500 steps (the first checkpoint) and remains 100\% throughout fine-tuning. Because in-distribution test set accuracy may not correlate with generalization accuracy, it is unclear which checkpoint would yield the best accuracy on the generalization set; we therefore report the mean generalization accuracy across all checkpoints.

\section{Architectural Effects}\label{sec:arch-effects}
Which architectural features contribute to hierarchical generalization? 
Given that language is structured hierarchically, we hypothesize that model depth (number of layers) will be the most important component, as deeper structure could more easily allow for representations of deeper hierarchical structures (e.g., more complex syntax trees), with recursive syntactic operations applied successively across layers \cite{murty-etal-2022-tree}. 

\begin{table}[t]
    \centering
    \resizebox{\linewidth}{!}{
    \begin{tabular}{lcccc}
    \toprule
    & Tiny & Mini & Small & Base \\
    \midrule
    \# Parameters & 16M & 31M & 60M & 220M \\
    \# Layers (NL) & 4 & 4 & 6 & 12 \\
    Feedforward layer dimension (FF) & 1024 & 1536 & 2048 & 3072 \\
    Embedding and hidden dimension (DM) & 256 & 384 & 512 & 768 \\
    Key/value projection matrix dim. (KV) & 32 & 32 & 32 & 64 \\
    \# Heads per layer (NH) & 4 & 8 & 8 & 12 \\
    \bottomrule
    \end{tabular}
    }
    \caption{Architectural details for T5 variants from \citet{tay-etal-2022-scale}. \# Layers is the number of layers in each of the encoder and the decoder (multiply this number by 2 to obtain the total number of layers in the model).}
    \label{tab:model-sizes}
\end{table}

\begin{figure}
    \centering
    \includegraphics[width=\linewidth]{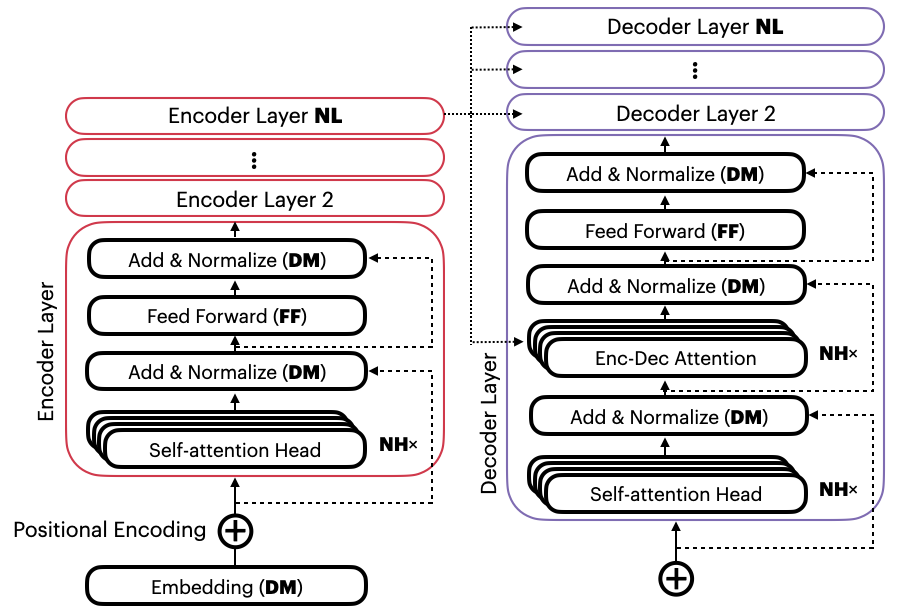}
    \caption{The Transformer encoder-decoder architecture, annotated with the architectural hyperparameters we vary.}
    \label{fig:arch-diagram}
\end{figure}

\subsection{Models}

We fine-tune pre-trained models from \citet{tay-etal-2022-scale}, available on HuggingFace. We train two sets of models. The first set is \texttt{google/t5-efficient-\{tiny,mini,small,}\textcolor{white}{,} \texttt{base\}}; see Table~\ref{tab:model-sizes} for the hyperparameters of these models, 
and Figure~\ref{fig:arch-diagram}  for a diagram of the Transformer architecture that illustrates these hyperparameters. Note that multiple hyperparameter values change at the same time when moving from, e.g., T5\textsubscript{small} to T5\textsubscript{base}. 

The second set of models we use from \citet{tay-etal-2022-scale} were derived from T5\textsubscript{base} by changing exactly one hyperparameter value. For these more controlled variants, we adopt \citeauthor{tay-etal-2022-scale}'s  nomenclature, which is based on the particular hyperparameter that is being changed, and its new value; for example, T5\textsubscript{base}-DM512 (which we abbreviate here to DM512) is identical to T5\textsubscript{base}, except the embedding/hidden dimension (DM) is reduced from 768 to 512. All of these models are trained on approximately 34B words from the Colossal Cleaned Common Crawl (C4) web text corpus.

\subsection{Depth, not Scale, Predicts Syntactic Bias}

\begin{figure}[t]
    \centering
    \includegraphics[width=0.82\linewidth]{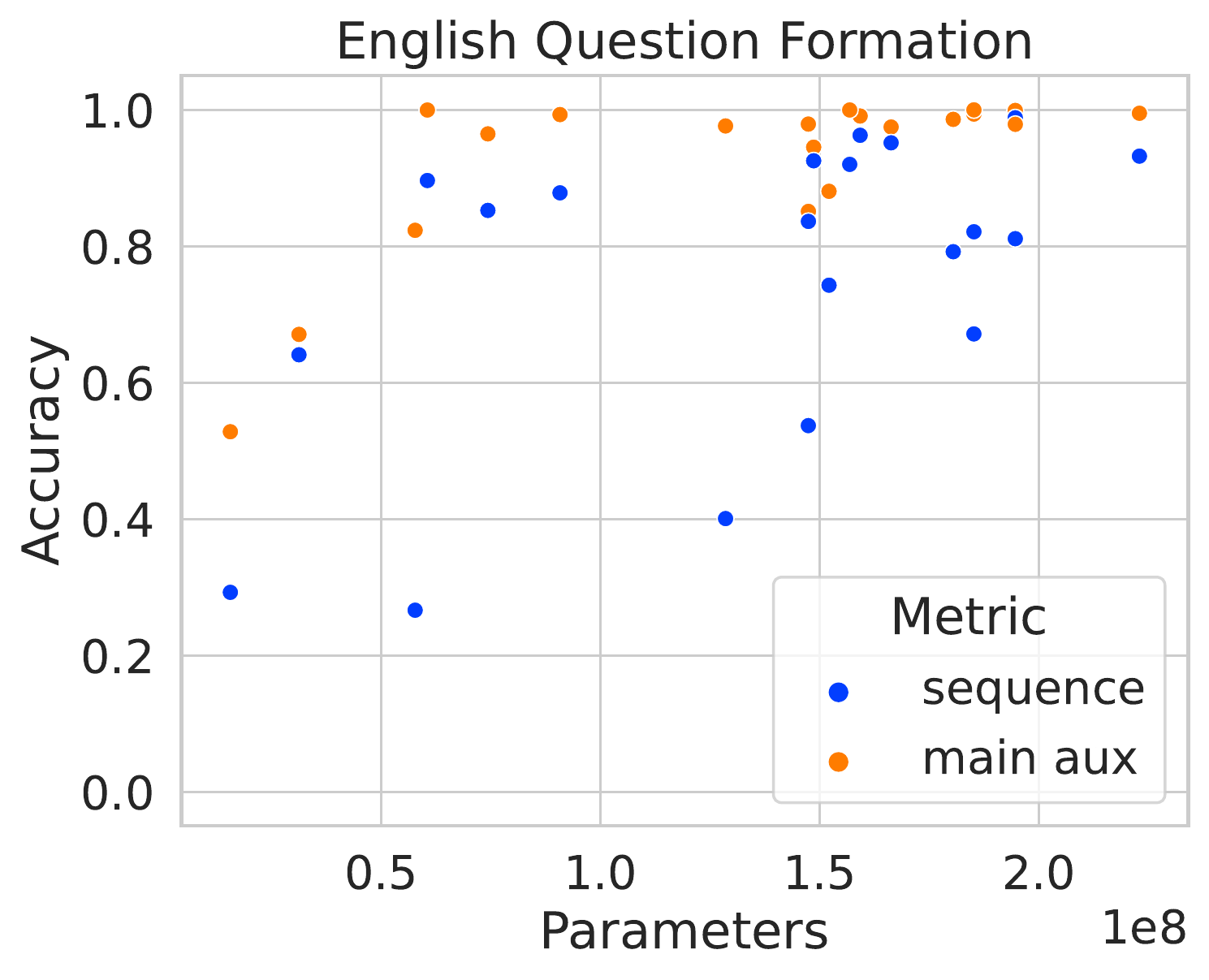}
    \includegraphics[width=0.82\linewidth]{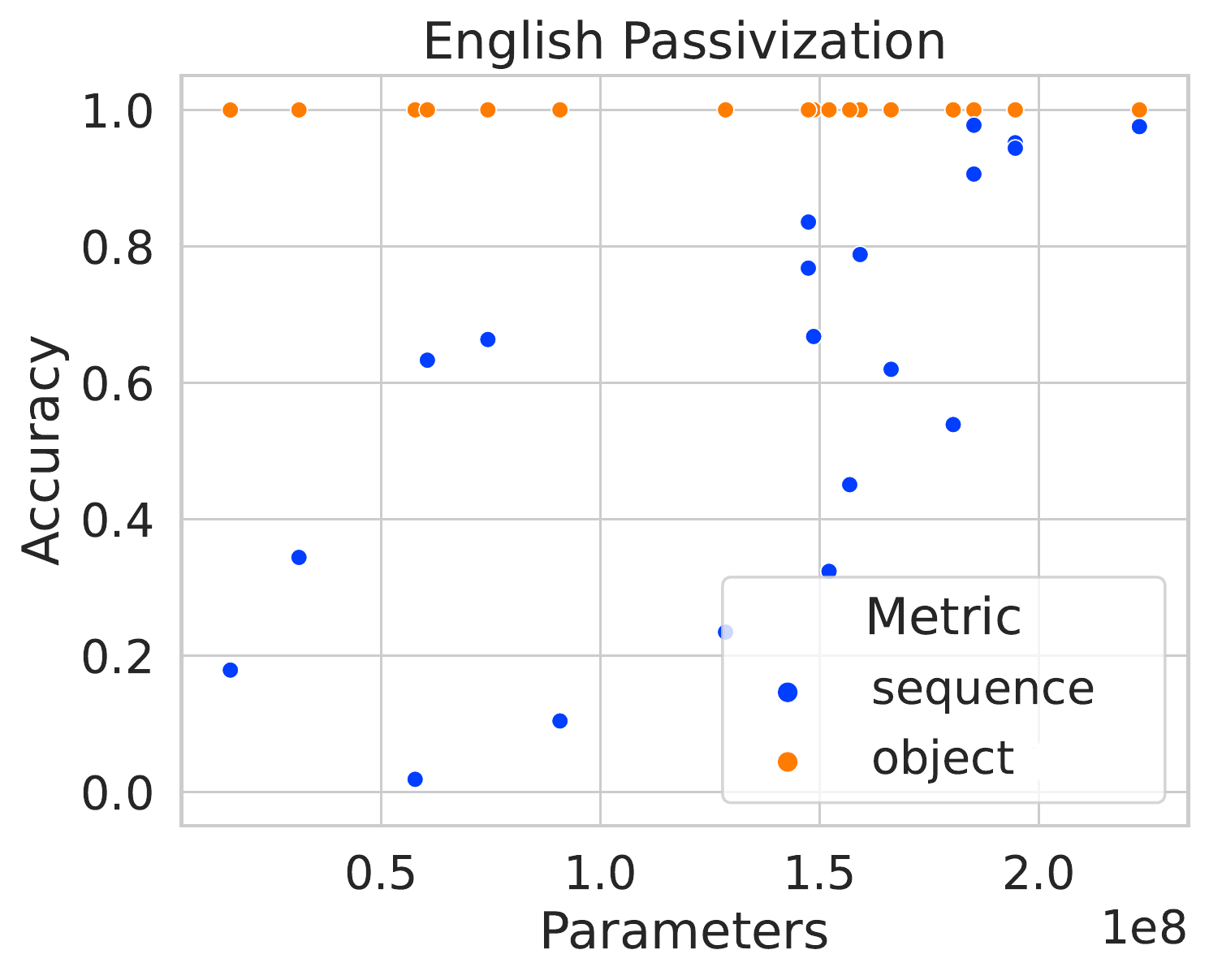}
    \caption{Generalization accuracies on question formation (top) and passivization (bottom) using T5\textsubscript{\{tiny,mini,small,base\}}, as well as variants of T5\textsubscript{base} where we vary the number of layers, number of attention heads per layer, embedding/hidden dimension, feed-forward width, or key-value projection dimension. There is a positive correlation between the number of parameters and accuracy, but the trend is not monotonic. Certain architectural features may therefore play a more important role than others.}
    \label{fig:parameterization}
\end{figure}

We start by asking whether scale alone can explain hierarchical generalization: Is there a monotonic relationship between the number of parameters and in generalization accuracy? We find that the answer is no (Figure~\ref{fig:parameterization}). For question formation, the Spearman rank-order correlation between the number of parameters and accuracy is 0.51 (sequence) and 0.58 (main auxiliary); for passivization, 0.75 (sequence) and 0.43 (object). While these are significant correlations ($p<.05$, except for object accuracy), if syntactic bias were predicted by scale alone, we would expect these to be close to~1. Thus, \textbf{number of parameters alone is not sufficient to explain the acquisition of a hierarchical bias}. This suggests that certain architectural components, which may be correlated with scale, are more important than others.

\begin{figure}[t]
    \centering
    \includegraphics[width=0.88\linewidth]{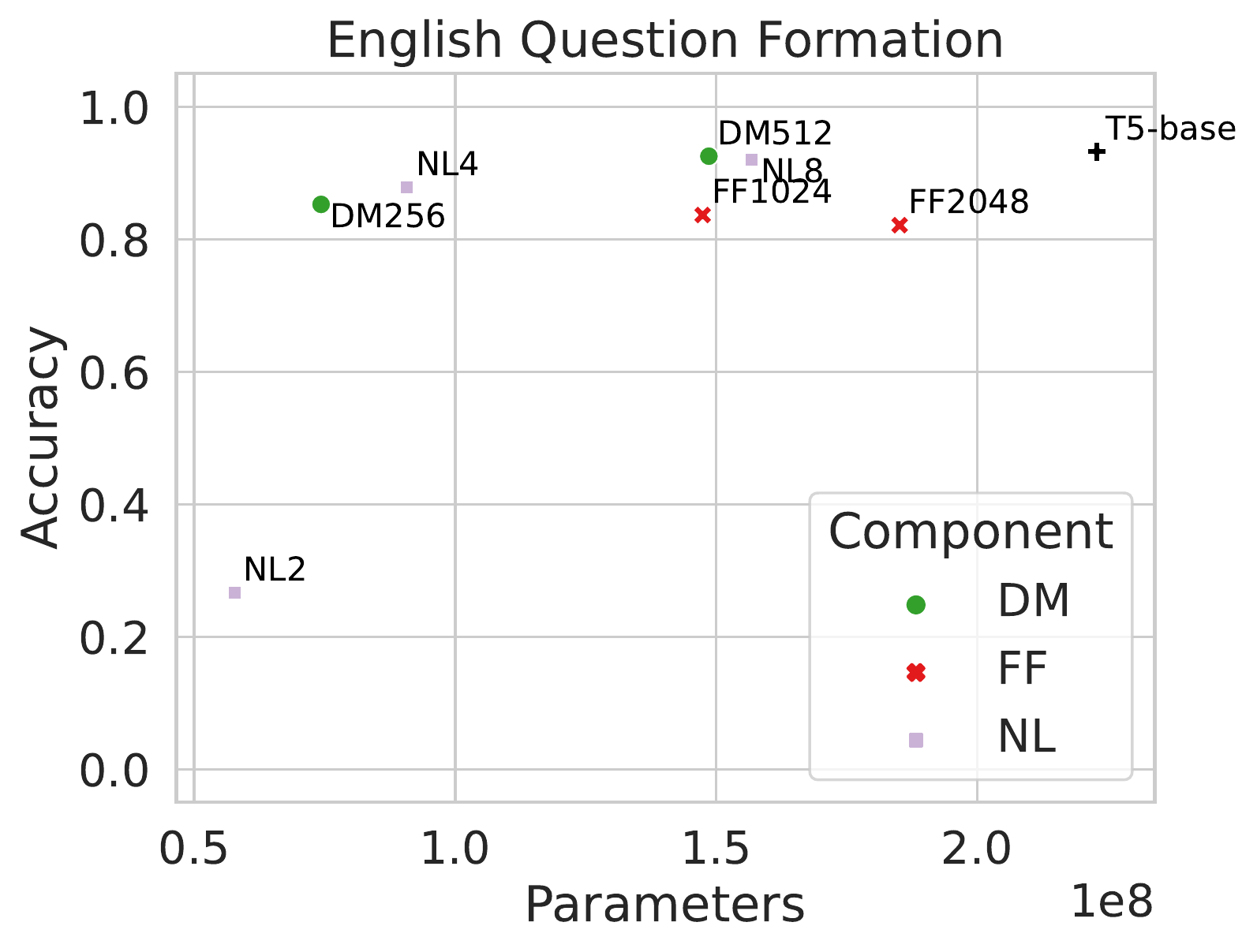}
    \includegraphics[width=0.88\linewidth]{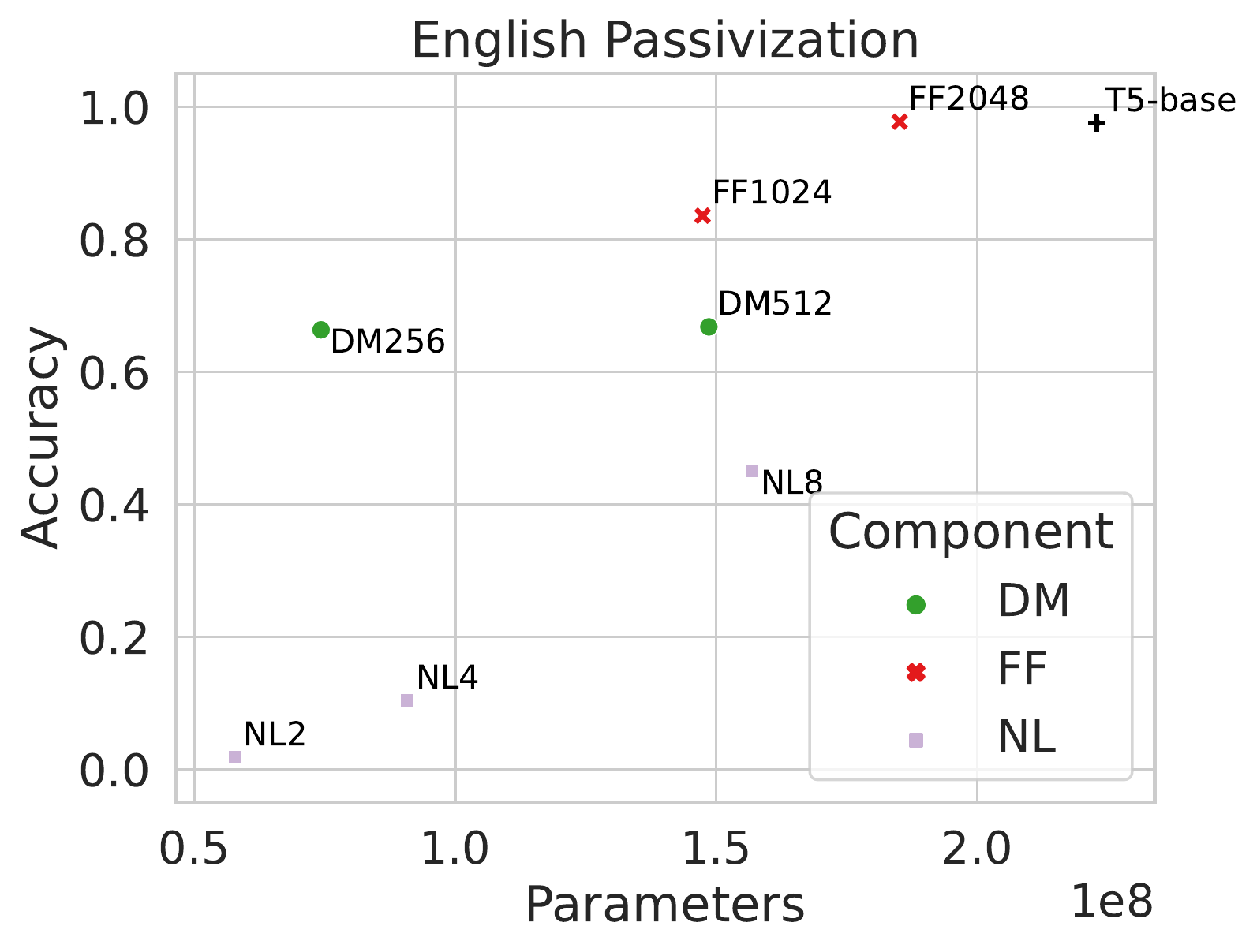}
    \caption{Generalization sequence accuracies on question formation (top) and passivization (bottom) using architecturally modified versions of T5\textsubscript{base}. Decreasing model depth (NL) results in the greatest drop in performance, suggesting that model depth is more important for learning syntax than other components. See App.~\ref{app:all-components} for results when varying the number of attention heads or key/value projection dimension.}
    \label{fig:arch-components}
\end{figure}

Indeed, we find that increasing model depth has a much stronger impact on accuracy than scaling the model up by increasing the value of other architectural hyperparameters (Figure~\ref{fig:arch-components}): in a least squares linear regression where the dependent variable is sequence accuracy and independent variable is number of parameters (normalized to the same range as the accuracy values), the slope of the fitted line is 0.70 when varying over number of layers, but only 0.13 for embedding/hidden size, and 0.25 for feed-forward layer width. In particular, the wide and shallow NL4 has more parameters than the narrow and deep DM256, but achieves similar performance as DM256 on question formation and significantly worse performance on passivization (as a reminder, NL4 is T5\textsubscript{base} with 4 encoder layers and 4 decoder layers, and DM256 is T5\textsubscript{base} with embedding/hidden size 256). This suggests that \textbf{when scaling the architecture, model depth is more important than other components for enabling hierarchical generalization}.

Is encoder depth or decoder depth more important for hierarchical generalization, or is total depth alone responsible for the patterns we find? We investigate this in App.~\ref{app:enc-dec}, with mixed results: for passivization, reducing the depth of either component leads to similar drops in generalization accuracy, but for question formation, decoder depth has a greater effect than encoder depth.

\subsection{Syntactic Bias Correlates with Downstream Performance}

How well does syntactic generalization accuracy correlate with performance on other tasks? We address this question by correlating main auxiliary accuracy with validation perplexity, question answering accuracy on SQuAD \citep{rajpurkar-etal-2016-squad}, and scores on the SuperGLUE collection of natural language understanding tasks \citep{wang-etal-2019-superglue}, all provided by \citet{tay-etal-2022-scale}. We do not report correlations with passivization accuracy, as most models achieve 100\% accuracy on this task, which leaves little explainable variance.

We obtain Spearman correlations of 0.57 ($p<$.1) for SuperGLUE, 0.34 ($p>$.1) for SQuAD, and 0.67 ($p<$.05) for negative validation perplexity. In other words, the correlation is weak but significant with average SuperGLUE accuracy (\citeauthor{tay-etal-2022-scale} do not report accuracy for individual SuperGLUE tasks); not significant with question answering; and relatively strong and significant with language modeling performance more broadly. We note that since the number of models is relatively modest, correlations need to be quite strong to reach the statistical significance threshold. 

These correlations do not indicate that syntactic abilities are \emph{causally} implicated in the models' improved performance on other tasks, but they do show that \textbf{the emergence of syntactic abilities often co-occurs with better language modeling performance and downstream performance}. Future work could employ causal analysis methods to better understand how the emergence of syntactic preferences affects (or does not affect) performance across NLP tasks.

\section{Corpus Genre}
\label{sec:style}
Large LMs are typically pre-trained on web text and/or Wikipedia data---genres that are distinct from the type of language that humans are exposed to during childhood. Could the domain of pre-training corpora explain why LMs require much more data than humans to reach similar syntactic abilities \cite{warstadt-etal-2020-learning}? \citet{huebner-etal-2021-babyberta} report experiments that support this hypothesis: they find that the RoBERTa masked LM achieves higher accuracies on linguistic acceptability judgment benchmarks when it is pre-trained on child-directed speech as opposed to a similar amount of Wikipedia data. In this section, we investigate whether this applies to our paradigm by pre-training encoder-decoder models on child-directed speech and a similar amount of text drawn from the English Wikipedia.

\subsection{Models}
We train models based on the T5 architecture and objective (see \S\ref{sec:methods}) on the English portion of CHILDES \citep{macwhinney2000childes}, a 5M-word child-directed speech corpus, and on an English Wikipedia corpus from \citet{huebner-etal-2021-babyberta}, which consists of a similar number of sentences as CHILDES. As Wikipedia sentences are longer, the total number of words in the Wikipedia training set we use here is approximately 10M.

We train models with eight hyperparameter configurations on each dataset (Table~\ref{tab:data-style}): we either vary the number of layers (NL $\in \{2,4,8,16\}$), keeping other hyperparameters, such as embedding/hidden dimension and number of heads, constant; or we keep the number of layers at 8 and vary other hyperparameters. While we only pre-train each configuration once, we fine-tune each configuration five times, with a different random seed each time.

Following \citeauthor{huebner-etal-2021-babyberta}, we modify the training hyperparameters to better suit the smaller and simpler child-directed speech corpus: we reduce the maximum sequence length to 128 and train a SentencePiece tokenizer \citep{kudo-richardson-2018-sentencepiece} with a reduced vocabulary size of $2^{13}=$ 8192; this is motivated by children's vocabulary size of approximately 5,000--6,000 lemmas at age~6 \citep{biemiller2003vocabulary}. For the Wikipedia corpus, we train SentencePiece tokenizers using vocab sizes $\in \{$8192, 32768$\}$ and take the best-performing model for each hyperparameter configuration,\footnote{32768 is the vocab size for T5 \citep{raffel-etal-2020-t5}.}$^,$\footnote{The best vocab size varied depending on model size and corpus size. Vocab size 8192 tends to work better for smaller corpora and smaller models on average, and 32768 tends to work better for larger corpora and larger models.} as it is not clear \emph{a priori} whether a smaller vocabulary would be beneficial for Wikipedia's more complex and diverse language. We use sequence packing, where we concatenate multiple sentences from the corpus into a single example such that the total length of each training example is approximately equal to the maximum sequence length. 

When pre-training on child-directed speech, we checkpoint every 10K training steps and find that the best performance on our syntactic transformations tasks is achieved at 130K steps. We train on the Wikipedia corpus for the same number of steps.

\begin{table}[t]
    \centering
    \resizebox{\linewidth}{!}{
    \begin{tabular}{lrrrrrrrr}
    \toprule
    & & & & & & & \multicolumn{2}{c}{\textbf{Data}}\\\cmidrule(lr){8-9}
    \textbf{Model} & Parameters & NL & FF & DM & KV & NH & CHILDES & Wikipedia \\
    \midrule
    Tiny & 23M & 8 & 1024 & 256 & 32 & 4 & 0.62 (.06) & 0.07 (.02) \\
    Mini & 50M & 8 & 1536 & 384 & 32 & 8 & 0.68 (.07) & 0.35 (.08) \\
    Small & 75M & 8 & 2048 & 512 & 32 & 8 & \textbf{0.73} (.04) & \textbf{0.46} (.10) \\
    Base & 157M & 8 & 3072 & 768 & 64 & 12 & 0.61 (.07) & 0.45 (.09) \\
    Large & 268M & 8 & 4096 & 1024 & 64 & 16 & 0.57 (.09) & 0.26 (.09) \\
    \midrule
    Small & 31M & 2 & 2048 & 512 & 32 & 8 & 0.49 (.04) & 0.08 (.01) \\
    Small & 46M & 4 & 2048 & 512 & 32 & 8 & 0.58 (.05) & 0.35 (.08) \\
    Small & 75M & 8 & 2048 & 512 & 32 & 8 & \textbf{0.73} (.04) & 0.46 (.10) \\
    Small & 134M & 16 & 2048 & 512 & 32 & 8 & 0.70 (.06) & \textbf{0.48} (.08) \\
    \bottomrule
    \end{tabular}}
    \caption{Main auxiliary accuracies averaged across 5 fine-tuning random seeds (standard deviation across seeds in parentheses) on the question formation generalization dataset for various encoder-decoder models pre-trained from scratch on 5M words of transcribed child-directed speech or 10M words of Wikipedia text.}
    \label{tab:data-style}
\end{table}

\subsection{Results}

We find that \textbf{pre-training on child-directed speech generally results in a greater ability to detect the main verb, as compared to pre-training on Wikipedia} (Table~\ref{tab:data-style}). This holds across model sizes and across model depths. The CHILDES-pre-trained 8-layer variant of T5\textsubscript{small} performs best. When fixing NL at 8 and varying other components according to each model size's default settings (as in Table~\ref{tab:model-sizes}), we find that T5\textsubscript{small} performs best. In the following experiment, we therefore focus on T5\textsubscript{small} modified to have 8 encoder layers and 8 decoder layers.

\section{How Much Data Leads to the Emergence of a Syntactic Bias?}\label{sec:data-quantity}
The next experiment we report has two goals. First, we aim to replicate the finding that simpler language gives rise to a stronger syntactic bias. Second, we expand the range of corpus sizes for the genres where larger corpora are available; our goal is to determine how much data is necessary to induce a hierarchical bias from each genre. In addition to child-directed speech and English Wikipedia, which we included in the previous experiment, we also pre-train models on the Colossal Cleaned Common Crawl (C4) web text corpus \cite{raffel-etal-2020-t5} and on Simple Wikipedia, which contains text from the same domain as English Wikipedia, but with a more limited vocabulary and simpler sentence structures.

\subsection{Method}

We collect English Wikipedia data using Wikidumps.\footnote{\url{https://dumps.wikimedia.org/enwiki/}} We use the \texttt{witokit} library\footnote{\url{https://github.com/akb89/witokit}} to preprocess the data. We pre-train on \{1M, 10M, 100M, 1B\} words of English Wikipedia data, where words are counted before being divided into subwords by the tokenizer. Our \{1M, 10M\}-word data is from \citet{huebner-etal-2021-babyberta}; our \{100M, 1B\}-word data is a concatenation of their 10M-word dataset with the Wikidump data that we download and preprocess. For C4, we randomly shuffle the HuggingFace version of the dataset\footnote{\url{https://huggingface.co/datasets/c4}} and sample individual examples until we have reached 1B words. We then create \{1M, 10M, 100M, 1B\}-word datasets by uniformly subsampling the data, ensuring that smaller datasets are subsamples of larger datasets. For CHILDES, we only have access to 5M words, so we pre-train on \{1M, 5M\} words, where the 1M-word dataset is a uniform subsample of the 5M-word dataset.

\begin{figure}[t]
    \centering
    \includegraphics[width=0.9\linewidth]{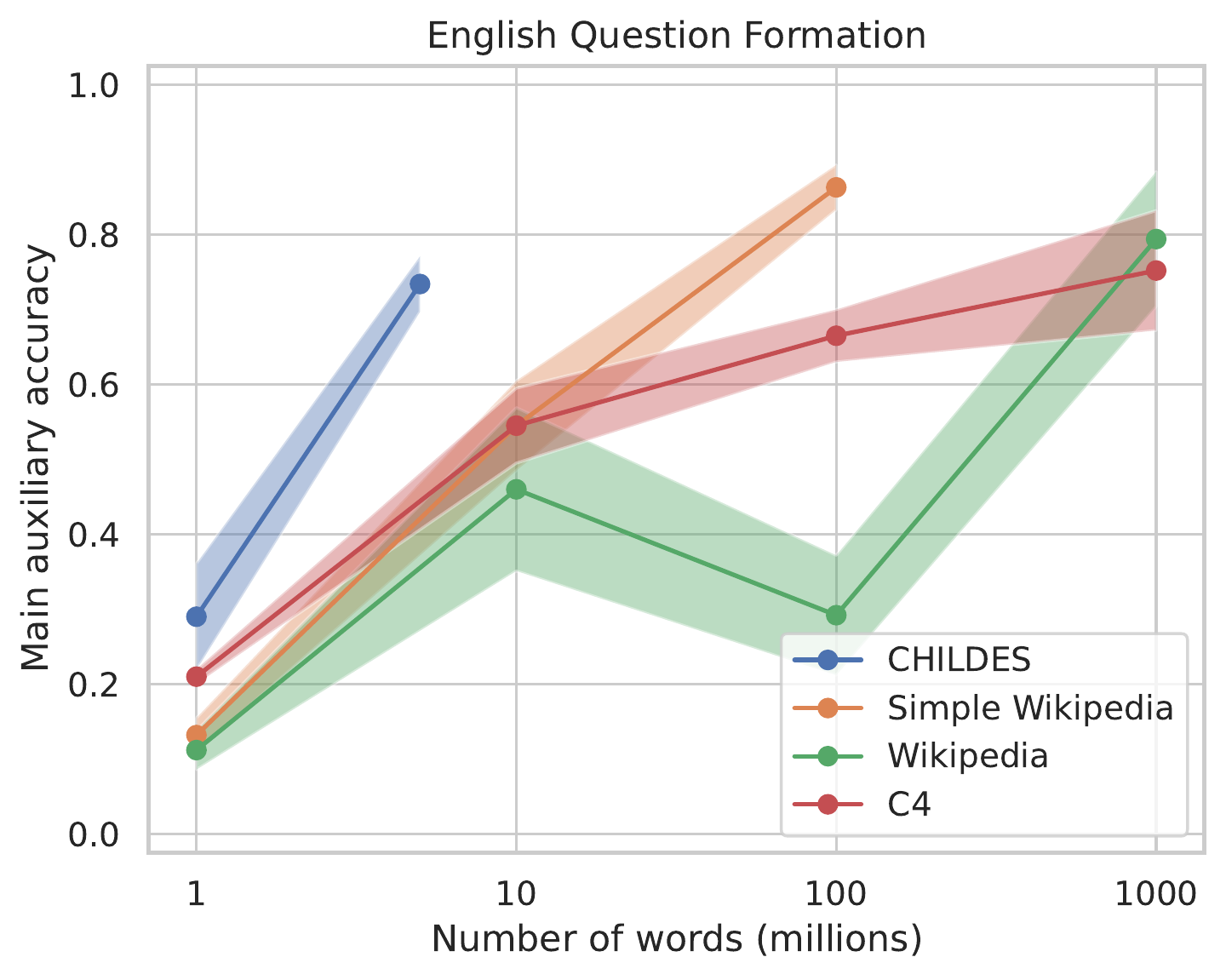}
    \includegraphics[width=0.9\linewidth]{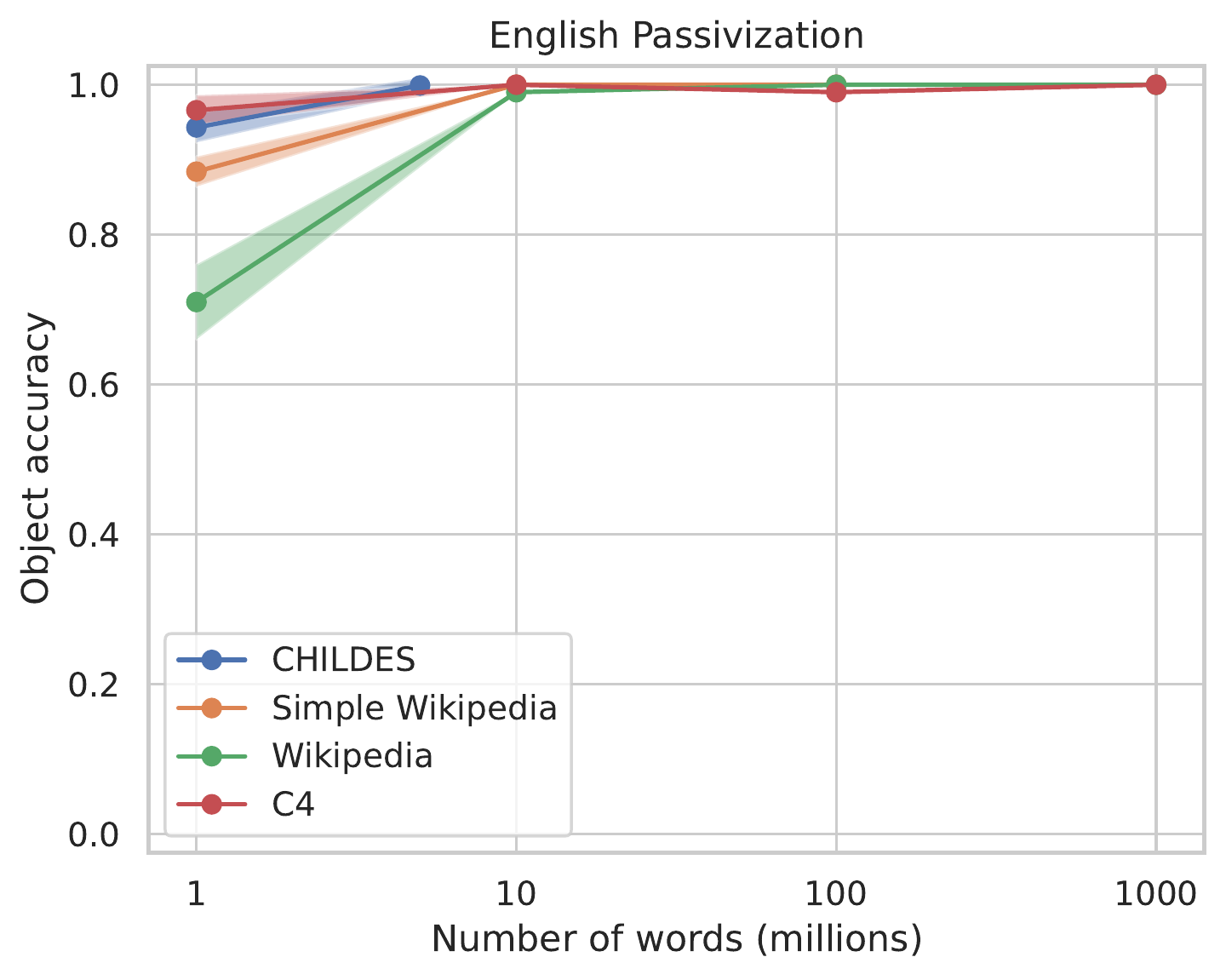}
    \caption{Generalization accuracies for question formation (top) and passivization (bottom) when pre-training a small T5-like model (8 encoder layers and 8 decoder layers) on corpora of various sizes and domains. Simpler language induces syntactic generalization with less data: CHILDES outperforms other datasets, and Simple Wikipedia outperforms Wikipedia. Accuracies (points) and standard deviations (shaded regions) are measured across 5 random seeds of fine-tuning.}
    \label{fig:data-scale}
\end{figure}

We also download Simple Wikipedia Wikidumps,\footnote{\url{https://dumps.wikimedia.org/simplewiki/}} and follow the same preprocessing pipeline we used for the English Wikipedia. Since we only have access to approximately 300M words of Simple Wikipedia, we only pre-train on \{1M, 10M, 100M\} words, where smaller datasets are uniform subsamples of larger datasets.

For all genres and sizes, we use the best-performing architecture from \S\ref{sec:style} (T5\textsubscript{small} with 8 encoder layers and 8 decoder layers), as well as the best training hyperparameters from that experiment. We tune over vocabulary size for each corpus style and size. See App.~\ref{app:hyperparameters} for details.

\subsection{Results}

Replicating and extending our results from \S\ref{sec:style}, we find that \textbf{pre-training on simpler language induces hierarchical generalization using less data} (Figure~\ref{fig:data-scale}). For question formation, transcribed child-directed speech, the simplest language style we use, induces hierarchical generalization in well over 50\% of question formation generalization examples using just 5M words. For Simple Wikipedia and C4, 100M words are required to reach this accuracy level; for Wikipedia, 1B words. Models pre-trained on Simple Wikipedia generalize in a much more syntax-sensitive manner than models pre-trained on a similar amount of Wikipedia data. 

For passivization, generalization accuracies are generally much higher, though the qualitative trends we observe for question formation still hold: child-directed speech induces hierarchical generalization using less data, and Simple Wikipedia induces hierarchical generalization using less data than Wikipedia.

\begin{table}[t]
    \centering
    \resizebox{\linewidth}{!}{
    \begin{tabular}{lccc}
        \toprule
         &  & Main auxiliary & Object \\
        Dataset & \# Words & accuracy & accuracy \\
        \midrule
        Wikipedia & 100M & 0.27 (.08) & 0.98 (.01) \\ 
        Wikipedia + CHILDES & 105M & 0.20 (.05) & 0.99 (.01) \\
        \midrule
        C4 & 100M & 0.67 (.04) & 1.00 (.00) \\
        C4 + CHILDES & 105M & 0.60 (.05) & 1.00 (.00) \\
        \midrule
        CHILDES & 5M & 0.73 (.04) & 0.99 (.00) \\
        \bottomrule
    \end{tabular}}
    \caption{Generalization accuracy  for question formation and passivization using the 100M-word versions of Wikipedia and C4, before and after concatenating CHILDES. Accuracy is averaged over five fine-tuning seeds (standard deviation over seeds in parentheses).}
    \label{tab:data-with-childes}
\end{table}

Could we narrow the gap between Wikipedia/C4 and CHILDES by simply concatenating CHILDES to these datasets? The answer appears to be no: performance does not significantly change when concatenating CHILDES to Wikipedia, nor when concatenating CHILDES to C4 (Table~\ref{tab:data-with-childes}). Perhaps the style of the different datasets is too dissimilar for the model to form consistent generalizations when exposed to both distributions simultaneously. It could be more beneficial to run a two-phase pre-training procedure, where we expose the model to the simpler CHILDES dataset first, and then expose it to Wikipedia or C4 only after it has acquired the hierarchical inductive bias. We discuss this hypothesis in more detail in \S\ref{sec:discussion}.

\section{Discussion}\label{sec:discussion}

\paragraph{Why does depth facilitate the emergence of a syntactic bias?}
Our first set of experiments suggests that depth is the most important architectural factor contributing to hierarchical generalization in Transformers. This finding is consistent with the suggestion of \citet{tay-etal-2022-scale}, who advocate for deeper and narrower architectures for the best performance across NLP tasks. Why are deeper models better in practice for many tasks and linguistic evaluations, when in theory an arbitrarily wide model can approximate any function with only two layers \citep{hornik-1989-universal}? 

One natural hypothesis is that Transformers generalize hierarchically on the basis of tree-structured representations organized across layers, such that higher layers represent larger constituents, and recursive syntactic operations are applied across successive layers; such a strategy arises more naturally in a deeper model. In recent work, \citet{murty-etal-2022-tree} find evidence that the internal organization of Transformer representations across layers becomes more tree-like over the course of training on some tasks, and that this property predicts the model's compositional generalization. While they fail to find a correlation between model depth and the degree to which representations are tree-shaped, this may be because they train relatively small models from scratch on synthetic 
datasets. In future work, methods such as those of \citet{murty-etal-2022-tree} may be used to measure the tree-likeness of Transformers' representations throughout pre-training on natural language, and the degree to which the tree-likeness of the pre-trained model correlates with the its syntactic inductive bias for fine-tuning.

\paragraph{Why does simpler language teach syntax more effectively?} We find that pre-training on simpler language, such as child-directed speech or Simple Wikipedia, enables hierarchical generalization from far less pre-training data than more complex language. Our findings from encoder-decoder models are consistent with previous findings from encoder-only masked LMs \cite{huebner-etal-2021-babyberta}, and with work on language understanding from speech \cite{gelderloos-etal-2020-learning}. The advantage of child-directed speech may be attributable to reduced lexical complexity, reduced syntactic complexity, or both \cite{soderstrom2007beyond}. Lower lexical complexity---in this case, fewer word types---may make it possible to learn the distribution of, say, parts of speech from a smaller corpus, as the same words would recur more often in different contexts. Lower syntactic complexity could result in a higher proportion of short sentences with unambiguous syntactic structure, which could help bootstrap syntactic learning. These two features are correlated in natural child-directed speech, but could be disentangled in future work by independently manipulating the lexical and syntactic distributions.

\paragraph{Simpler language can be leveraged for more efficient pre-training.} Our experiments show that not all pre-training data is created equal, and motivate further research on data curation for pre-training, and in particular on curriculum learning \cite{bengio-2009-curriculum}. We conjecture that robust syntactic inductive biases will play a role not only in fine-tuning but also in pre-training, making it possible for models to use additional pre-training sentences more efficiently. This motivates a two-phase ``starting small'' approach \citep{elman-1993-learning}, where the model is first exposed a model to child-directed speech until syntactic inductive biases emerge, and then pre-training on a larger corpus proceeds as usual afterwards. This approach is related to, but distinct from, the single-phase simple-to-complex approach, where a pre-training dataset is sorted from the simplest inputs to the most complex and then presented to a model in order. The single-phase approach has demonstrated mixed results \citep{campos-2021-curriculum,surkov-etal-2022-data}, but to our knowledge, a syntax-focused two-phase approach has not yet been attempted.

\paragraph{Transformers may be more data-efficient than previously thought.} Our findings about the amount of pre-training data required for the acquisition of syntactic biases also have implications for cognitive modeling research. Humans learn language from far fewer words than contemporary LMs, and at the same time generalize their linguistic knowledge to new settings more robustly; conversely, standard NLP evaluations, which do not take the pre-training corpus into consideration, implicitly reward architectures that learn well from vast amounts of data, raising the concern that those architectures are suboptimal for cognitive modeling \citep{linzen-2020-accelerate}. Our evaluation setup and empirical results go some way towards addressing these concerns: we show that pre-training on a developmentally plausible amount of data can induce human-like inductive biases that improve out-of-distribution generalization. This suggests that Transformers, when trained in cognitively relevant regimes, may serve as fruitful models of human language acquisition and processing (see also \citealt{hosseini2022artificial}).

\section{Conclusions}
We have analyzed the architectural and data features that contribute to the acquisition of syntactic inductive biases during the pre-training of encoder-decoder Transformers. We find that model depth matters more for hierarchical generalization than other model components (\S\ref{sec:arch-effects}); that models more quickly learn that language is hierarchical given simpler language (\S\ref{sec:style}); and that it takes orders-of-magnitude more data to induce hierarchical inductive biases when pre-training on genres such as Wikipedia or web text, compared to simpler data such as child-directed speech (\S\ref{sec:data-quantity}).

\section*{Acknowledgements}

We thank the authors of \citet{tay-etal-2022-scale} for facilitating academic research on language model scaling by releasing a large range of model checkpoints. We also thank Alexandra DeLucia, Nathaniel Weir, and Daniel Khashabi for their thoughtful feedback on earlier versions of this paper. This material is based upon work supported by the National Science Foundation (NSF) under Grant No. BCS-2114505. Aaron Mueller was supported by a National Science Foundation Graduate Research Fellowship (Grant \#1746891). This work was supported in part through the NYU IT High Performance Computing resources, services, and staff expertise.

\section*{Limitations}
Our analyses are based on models with T5-like architectures and span denoising training objectives. Thus, our findings may not generalize to other types of encoder-decoder models (e.g., BART), nor encoder-only and decoder-only models. We believe this is unlikely, given that similar findings have been shown for models with architectures and objectives that differ significantly from T5's \cite{huebner-etal-2021-babyberta,warstadt-bowman-2020-linguistic}. Nonetheless, it cannot be ruled out.

Our analyses are also based entirely in English, and only leverage two syntactic transformations. It is possible that our findings will not generalize to other languages, given that certain grammatical features (e.g., more extensive case marking) induce more syntax-sensitive behavior given a similar amount of training data across languages \citep{mueller-etal-2020-clams,ravfogel-etal-2019-studying}; thus, perhaps less Wikipedia or C4 data is needed in these languages for models to acquire hierarchical preferences. It is also possible that, within a language, a model could adopt a hierarchical inductive bias for one type of transformation, but not another---especially if one transformation is much more frequent than the other. Indeed, the frequency of particular words positively correlates with syntactic evaluation accuracies \citep{wei-etal-2021-frequency,newman-etal-2021-refining}, and it would be reasonable to expect a similar trend for the frequency of syntactic transformations. Thus, future work should investigate more transformations in more languages to ensure that these findings are consistent.

\bibliography{custom}
\bibliographystyle{acl_natbib}

\newpage
\appendix

\section{Hyperparameters}\label{app:hyperparameters}
When fine-tuning models on syntactic transformations, we use settings from \citet{mueller-etal-2022-coloring}: batch size 128, window size 128, initial learning rate of $5\times10^{-5}$, fine-tune for 10 epochs ($\approx$7500 training steps), checkpoint and evaluate every 500 steps.

When pre-training models from scratch, we train for 130K training steps, batch size 16 (except for the 1B-word datasets, where we use batch size 128 such that the model sees the entire dataset at least once). We tune over the vocabulary size $\in$ \{8192, 32768\} for each dataset and dataset size.

\section{Is the Encoder or Decoder More Important for Hierarchical Generalization?}
\label{app:enc-dec}

In \S\ref{sec:arch-effects}, we found that model depth is more important than model width for enabling LMs to acquire a hierarchical inductive bias. Here, we specifically investigate whether the encoder or decoder of the model is more important by varying the depth of the encoder and decoder individually and observing changes in generalization patterns. As in \S\ref{sec:arch-effects}, our models are based on the T5\textsubscript{base} architecture, which has 12 encoder and 12 decoder layers.

\begin{figure}[t]
    \centering
    \includegraphics[width=0.9\linewidth]{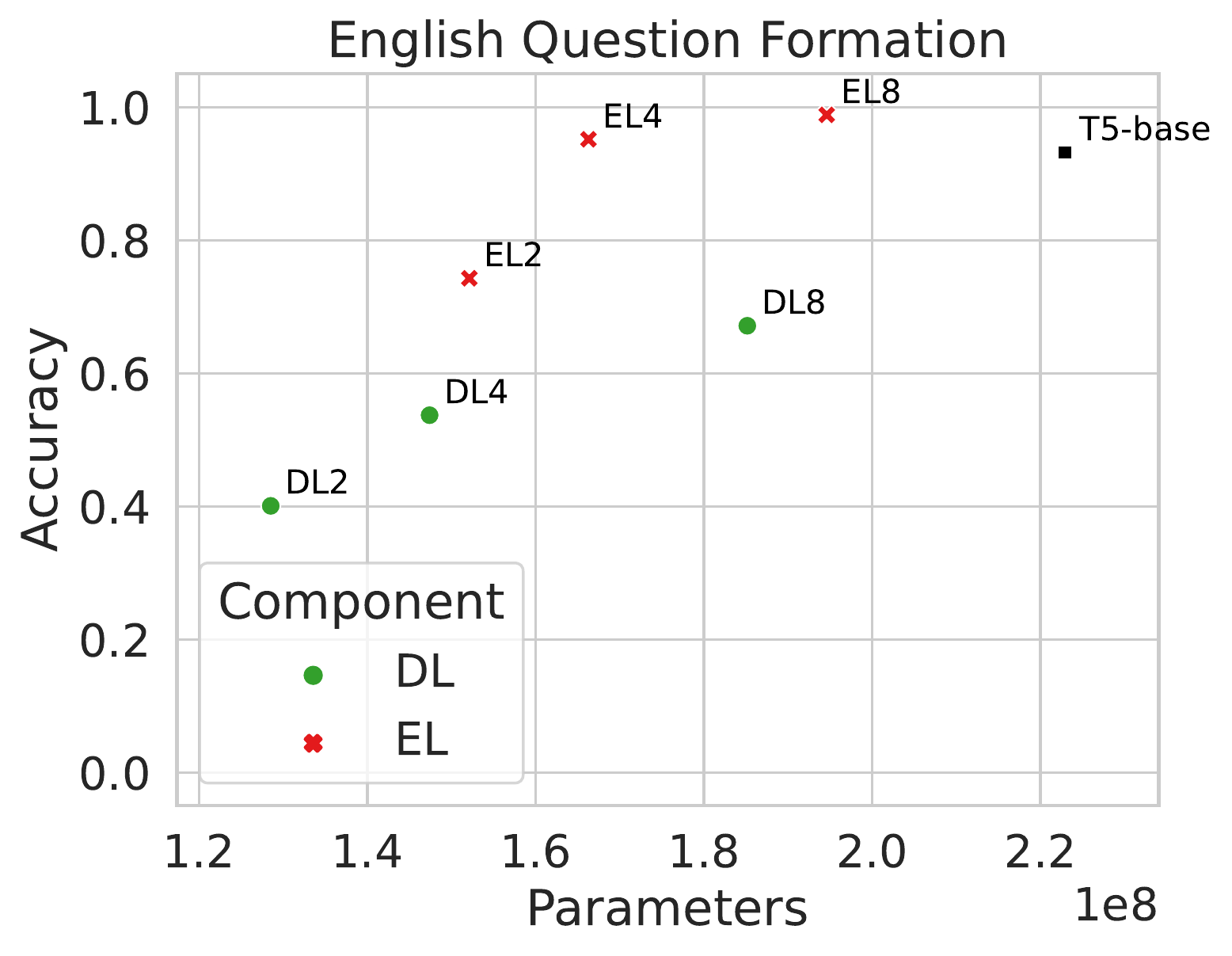}
    \includegraphics[width=0.9\linewidth]{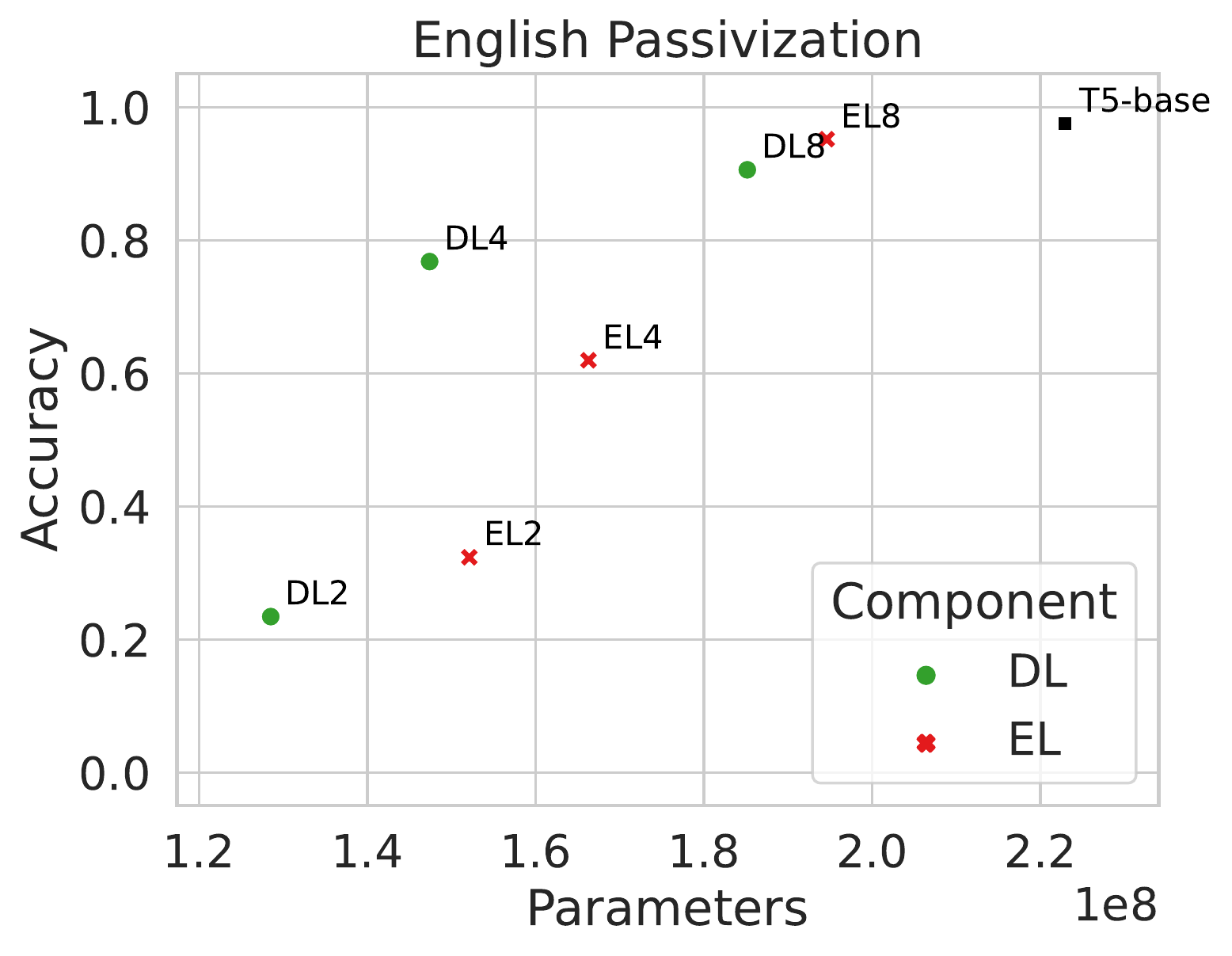}
    \caption{Generalization sequence accuracies on question formation (top) and passivization (bottom) using T5\textsubscript{base}, as well as variants of T5\textsubscript{base} where we vary the encoder depth/decoder depth. Here, EL2 refers to an architecture identical to T5\textsubscript{base}, except it has 2 encoder layers (instead of 12). Likewise, DL2 is created from T5\textsubscript{base} by modifying the number of decoder layers, keeping the number of encoder layers at the original 12.}
    \label{fig:eldl-components}
\end{figure}

In our results (Figure~\ref{fig:eldl-components}), we observe that decreasing the depth of either component leads to similar losses in accuracy on passivization, though decreasing decoder depth results in consistently lower accuracies for question formation. Thus, total depth may be the most important factor, regardless of where it is concentrated. Nonetheless, we observe preliminary evidence for the decoder being slightly more important for acquiring a hierarchical inductive bias---or at least generating outputs that are consistent with this bias for question formation. Future work could investigate other transformations and other languages to test the consistency of these findings.

\section{All Architectural Variation Results}\label{app:all-components}
\begin{figure}[t]
    \centering
    \includegraphics[width=0.9\linewidth]{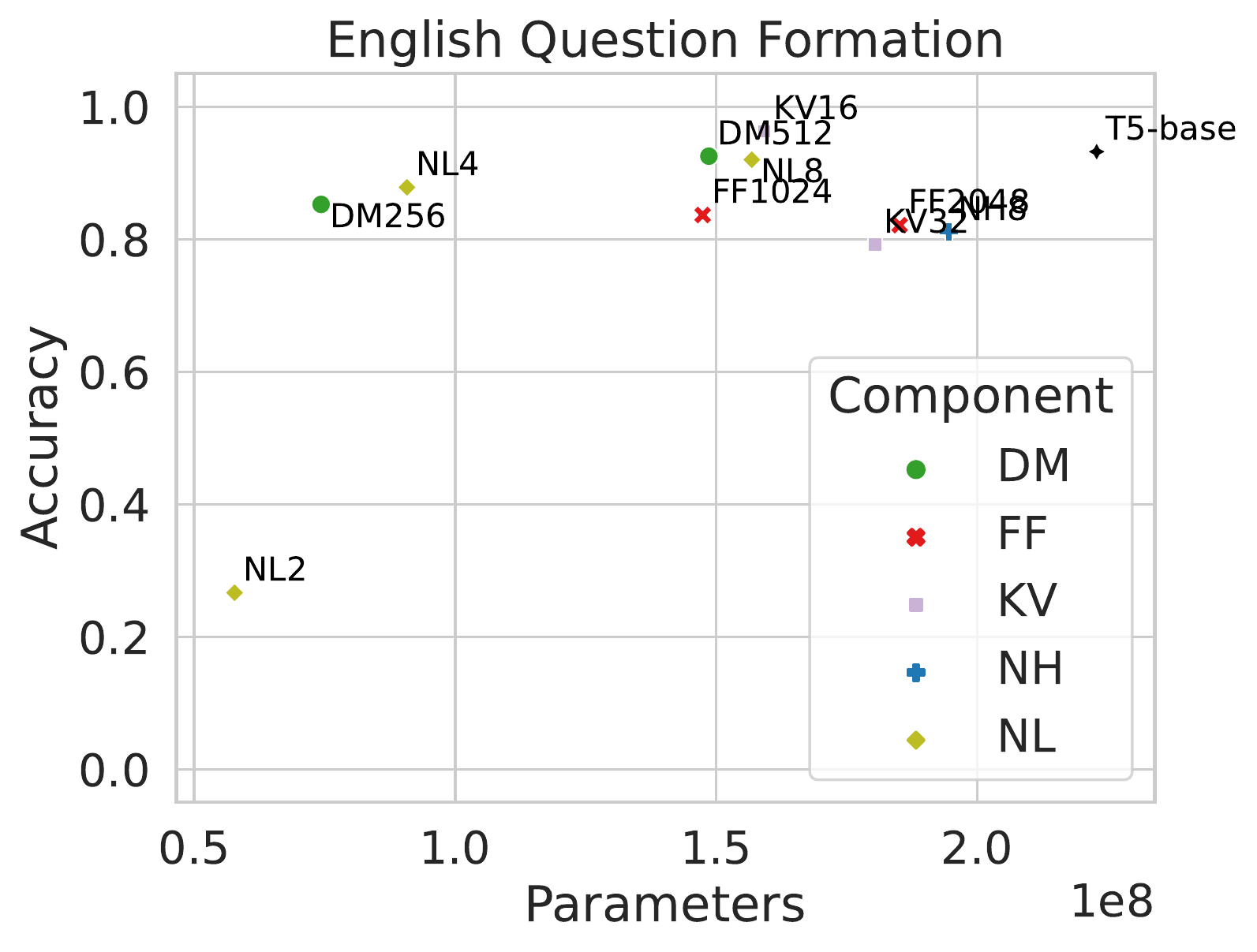}
    \includegraphics[width=0.9\linewidth]{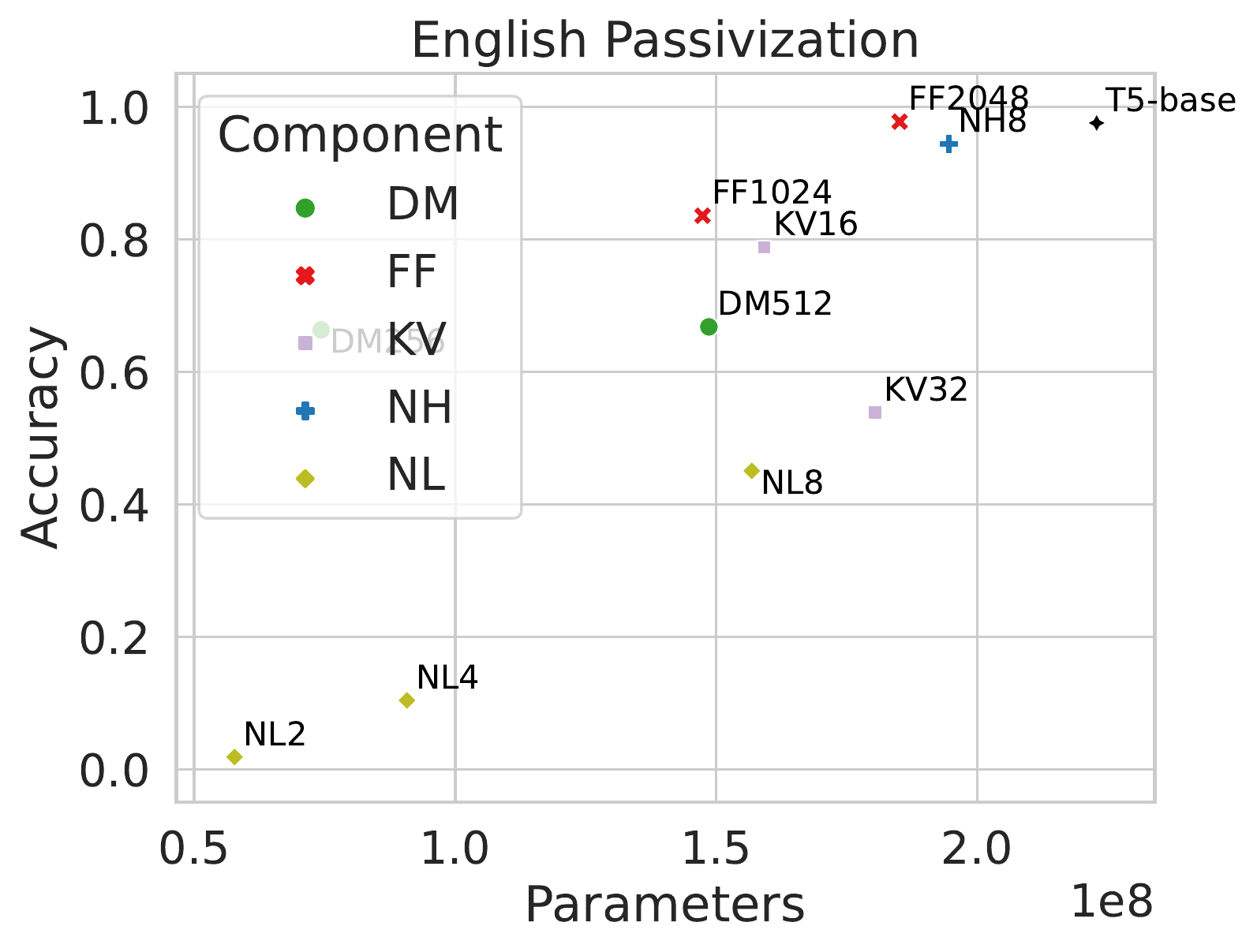}
    \caption{Generalization sequence accuracies on question formation (top) and passivization (bottom) using all architectural variants of T5\textsubscript{base}, as well as T5\textsubscript{base} itself.}
    \label{fig:all-components}
\end{figure}

In \S\ref{sec:arch-effects} and App.\ \ref{app:enc-dec}, we show that model depth is more important than model width. However, we did not show the performance of models where we vary the number of attention heads, nor the key-value projection matrix dimension. Here, we show the full results (Figure~\ref{fig:all-components}).

Overall, varying the number of attention heads has little effect on the performance of the model. We see the same trend for reductions in the size of the key/value projection matrix. Thus, model depth still appears to be the most important component in inducing hierarchy-sensitive generalizations.

\end{document}